\documentclass[letterpaper]{article} 
\usepackage{aaai2027}
\usepackage[hyphens]{url}  
\usepackage{graphicx} 
\usepackage{natbib}  
\usepackage{caption} 
\usepackage{algorithm}
\usepackage{algorithmic}
\usepackage{amsmath}
\usepackage{amssymb}

\usepackage{siunitx}
\usepackage[colorlinks=true, urlcolor=blue, linkcolor=black, citecolor=black]{hyperref}

\usepackage{newfloat}
\usepackage{listings}
\DeclareCaptionStyle{ruled}{labelfont=normalfont,labelsep=colon,strut=off} 
\floatstyle{ruled}
\newfloat{listing}{tb}{lst}{}
\floatname{listing}{Listing}

\usepackage{booktabs}
\usepackage{multirow}

\usepackage{marvosym}

\nocopyright 

\title{DreamWAM: Beyond RGB Future Prediction for World Action Models}
\author{
    Shanglin Yuan\textsuperscript{\rm 1,\rm 2,\ensuremath{*}},
    Weiheng Zhao\textsuperscript{\rm 1,\rm 2,\ensuremath{*}},
    Xin Shi\textsuperscript{\rm 2},
    Haoyi Jiang\textsuperscript{\rm 1,\rm 2},
    Xianda Guo\textsuperscript{\rm 3},\\
    Liu Liu\textsuperscript{\rm 4},
    Wenyu Liu\textsuperscript{\rm 1},
    Wei Sui\textsuperscript{\rm 2,\ensuremath{\dagger}} ,
    Xinggang Wang\textsuperscript{\rm 1,\ensuremath{\ddagger}}
}
\affiliations{
    \textsuperscript{\rm 1}Huazhong University of Science and Technology \quad
    \textsuperscript{\rm 2}D-Robotics  \quad
    \textsuperscript{\rm 3}Wuhan University \quad
    \textsuperscript{\rm 4}Horizon Robotics \\
    \textsuperscript{\ensuremath{*}}~Equal contribution\quad
    \textsuperscript{\ensuremath{\dagger}}~Project Lead\quad
    \textsuperscript{\ensuremath{\ddagger}}~Corresponding Author
}

\begin{document}

\maketitle

\begin{abstract}
World Action Models (WAMs) learn action-relevant representations by predicting how the observed world will evolve.
Most existing WAMs define this future in RGB space, where task-relevant state transitions are entangled with nuisance variations in texture, illumination, background, and viewpoint.
We argue that WAMs should explicitly predict action-relevant future state rather than relying on RGB prediction alone.
We introduce DreamWAM, which reformulates future prediction as structured world modeling beyond RGB, representing future states through complementary views of appearance, motion, geometry, and semantics.
During training, DreamWAM combines joint latent denoising of RGB and motion with lightweight gated residual branches for geometry and semantics.
Shared attention between VideoDiT and ActionDiT allows the action branch to learn from these future-state predictions, while all beyond-RGB supervision branches are disabled at inference and deployment remains RGB-only.
Across both no-rollout and joint video-action inference, DreamWAM consistently improves the matched RGB-only baselines on LIBERO, from 97.30\% to 98.40\% and from 98.00\% to 98.90\%, respectively.
The gains become larger under unseen LIBERO-Plus perturbations, from 51.36\% to 63.44\% and from 69.16\% to 75.47\%.
The same robustness extends to real-world manipulation, where DreamWAM attains an average success rate of 74.4\% across unseen changes in lighting, background, and object layout, compared with 55.6\% for Fast-WAM-Joint.
These results show that robust world-action learning depends not only on predicting the future, but on representing it in a form that matters for action. The code and models are publicly released at \url{https://github.com/hustvl/DreamWAM}.
\end{abstract}

\section{Introduction}
\label{sec:introduction}

World Action Models (WAMs) learn to act by coupling future prediction with
action generation
\citep{worldvla,motus,dreamzero,cosmospolicy,lingbot-va,fastwam,wav}. Rather than
mapping the current observation directly to an action, a WAM anticipates how an
interaction may unfold and exposes the resulting world representation to an
action expert. Future prediction therefore serves not only as a rollout
mechanism, but also as an inductive bias for learning scene dynamics,
interaction progress, and task completion. Recent WAMs have primarily advanced
\emph{how} this future is predicted through stronger video backbones, tighter
world-action coupling, and more efficient generation
\citep{dreamzero,cosmospolicy,lingbot-va,x-wam,fastwam}. A more basic
question, however, remains underexplored: \emph{what should a WAM imagine about
the future?}

\begin{figure*}[t]
    \centering
    \includegraphics[
        width=\textwidth,
    ]{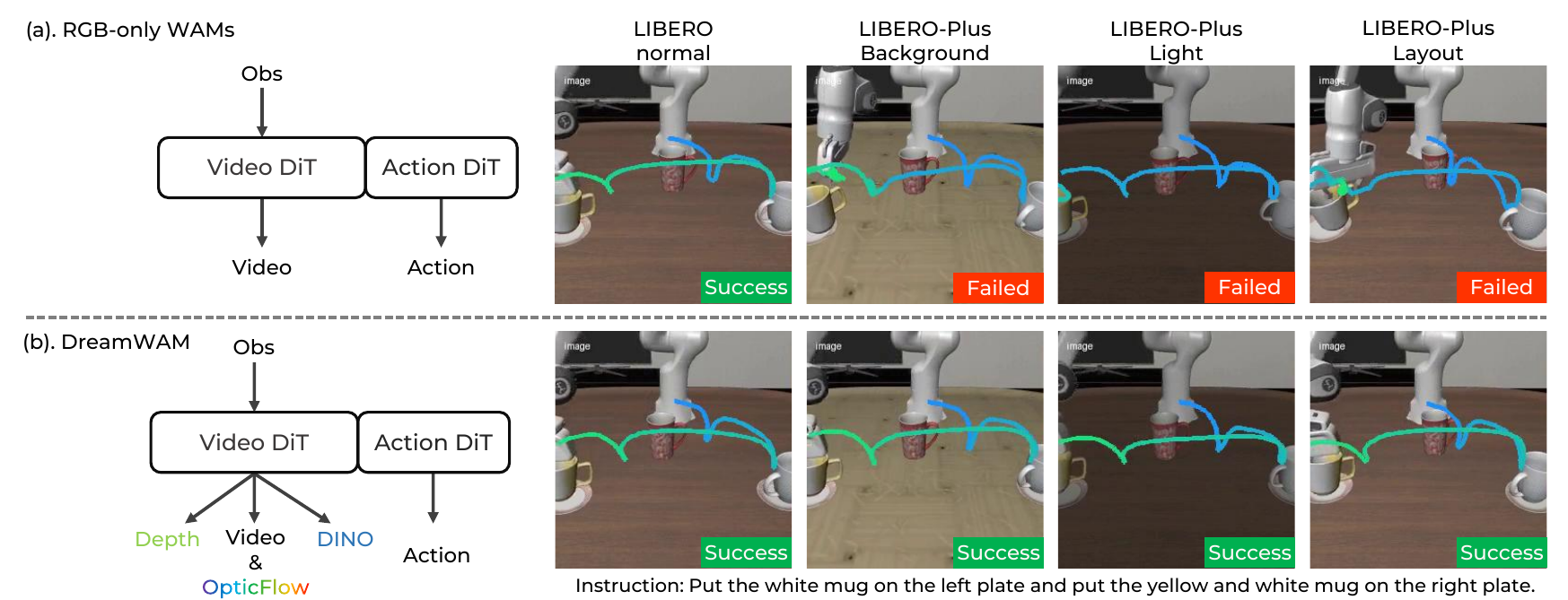}
    \caption{\textbf{Dreaming beyond RGB.} RGB-only WAMs organize future
    learning around video appearance, whereas DreamWAM additionally learns
    motion, geometry, and semantic views of future state. Representative
    LIBERO rollouts under the same instruction show that the RGB-only baseline
    succeeds in the original scene but fails under unseen background,
    dimmed lighting, and layout changes, while DreamWAM completes the task across all
    four settings.}
    \label{fig:teaser}
\end{figure*}

A useful anticipation of an interaction is more than a pixel-perfect rendering
of its next frame. Consider the manipulation in
\autoref{fig:teaser}: placing two mugs on their corresponding plates requires
the policy to preserve which mug is which, anticipate how each mug should move,
and reason about where it should end relative to the plates. When we imagine or
dream about such an interaction, its useful content does not depend on
reproducing every color, texture, or background exactly; what must remain
coherent is \emph{what} is involved, \emph{how} it changes, and \emph{where}
that change occurs. This suggests a richer notion of future imagination for
WAMs: a machine ``dream'' should represent not only what the future looks like,
but also its dynamics, spatial organization, and semantic structure.

Most existing WAMs instead define the future almost exclusively in RGB
observation space. RGB contains motion, geometry, and semantic cues, but an
RGB-only prediction objective entangles task-relevant state transitions---such
as object displacement, contact, and goal completion---with variations in
texture, illumination, background, and viewpoint that need not change the
required action. As illustrated in \autoref{fig:teaser}, this distinction is
particularly important under visual distribution shifts: the manipulation
objective can remain unchanged even when the observed pixels differ
substantially. A future representation organized only around appearance may
therefore fit visually salient details without explicitly preserving the state
evolution that determines action success.

Built on this view, we introduce \textbf{DreamWAM}, which reformulates future
prediction as structured world modeling beyond RGB. DreamWAM describes future
states through four complementary views: appearance captures what the future
looks like; motion makes temporal change explicit; geometry characterizes
spatial organization; and semantics preserves object- and region-level
consistency. These views are correlated projections of action-relevant future
state rather than independent factors or a complete physical description. To
respect their heterogeneous forms, DreamWAM jointly denoises RGB and
RAFT-derived optical-flow latents~\citep{raft} during training, while lightweight
gated residual branches model future geometry and semantics using targets from
Depth Anything V3~\citep{depthanythingv3} and DINOv2~\citep{dinov2}. This
residual design retains the pretrained video pathway as the primary
representation stream and applies the feature-level supervision through
controlled corrections. Shared attention between VideoDiT and ActionDiT allows
the resulting future-state representation to directly shape action learning.

The richer future representation is learned without altering policy deployment.
All non-RGB supervision pathways are inactive at test time: the motion input is
disabled, while the external target encoders, and
feature-prediction heads are removed. Inference therefore follows the RGB-only deployment interfaces of Fast-WAM and outputs only the denoised action chunk. DreamWAM changes what the policy
learns to imagine while preserving the deployment pathway of the underlying
WAM.

We evaluate DreamWAM in simulation and on a real robot. Across matched
no-rollout and joint settings, DreamWAM raises average success on LIBERO from
97.30\% to 98.40\% and from 98.00\% to 98.90\%, respectively. Under unseen
LIBERO-Plus perturbations, the corresponding averages increase from 51.36\%
to 63.44\% and from 69.16\% to 75.47\%, with improvements across all seven
shifts in both settings. The same pattern holds in real-world manipulation:
DreamWAM improves the standard-task average from 90.8\% to 96.7\% and attains
74.4\% success across unseen changes in lighting, background, and object
layout, compared with 55.6\% for Fast-WAM-Joint. The consistent gains across
inference modes, amplified under visual shifts, support our central claim that
robust world-action learning depends not only on predicting the future, but on
representing it in a form that preserves what matters for action.

Our main contributions are threefold:
\begin{itemize}
    \item We identify the representation of the predicted future as a core
    design variable in WAMs and formulate future imagination beyond RGB through
    complementary appearance, motion, geometry, and semantic views.
    \item We introduce DreamWAM, which combines RGB-motion joint latent
    denoising with lightweight gated residual modeling of future geometry and
    semantics during training, while reverting to the RGB-only Fast-WAM
    VideoDiT--ActionDiT pathway at deployment.
    \item We validate DreamWAM through matched simulation and real-world
    evaluations. It improves performance under both no-rollout and joint
    inference, with substantially larger gains under unseen visual perturbations.
\end{itemize}

\section{Related Work}
\label{sec:related_work}

\subsection{Vision--Language--Action Models and World Action Models}

Vision--language--action (VLA) models map visual observations and language
instructions to robot actions. RT-1, RT-2, and Octo established scalable
generalist policies, while OpenVLA, UniVLA, OpenVLA-OFT, CogACT, the
$\pi$-series, X-VLA, and SmolVLA improve action modeling, cross-embodiment
transfer, and efficient adaptation
\citep{rt1,rt2,octo,openvla,univla,openvla-oft,cogact,pi0,pi0-fast,
pi05,pi06,pi07,xvla,smolvla}. Foundation-scale systems such as
LingBot-VLA and Qwen-VLA further broaden data and embodiment coverage
\citep{lingbotvla,qwenvla}. These policies generally predict actions
directly without explicitly representing future world evolution.

Predictive robot policies instead use video or latent world modeling to
structure control. UniPi generates goal-conditioned future videos before
recovering actions, whereas recent predictive policies and WAMs jointly or
tightly couple future-state
and action prediction through autoregressive, diffusion, or shared-expert
architectures
\citep{unipi,uwm,worldvla,motus,dreamzero,cosmospolicy,bagelvla,
lingbot-va,x-wam,gigaworld-policy,drivevla-w0,fastwam,wav}. Related predictive models also study efficient long-horizon generation \citep{sana-wm}. Most existing WAMs nevertheless organize the predicted
future primarily in RGB or generic video space. DreamWAM complements this line
by studying the representation of the future itself and modeling appearance,
motion, geometry, and semantics as complementary signals for action learning.

\subsection{Auxiliary World Knowledge in VLA and Video Models}

In visual generation, REPA aligns DiT hidden states with pretrained visual
representations, while VA-VAE applies analogous alignment to VAE latents
\citep{repa,vavae}.

A growing line of work injects auxiliary world knowledge into VLA
representations. Spatial Forcing, DepthVLA, SpatialVLA, GeoVLA, PointVLA, and
BridgeVLA use depth, point clouds, 3D-aligned encodings, or spatial-feature
alignment to strengthen geometric reasoning
\citep{spatialforcing,depthvla,spatialvla,geovla,pointvla,bridgevla}.
PointWorld extends structured prediction to action-conditioned 3D point-flow
dynamics \citep{pointworld}. ReconVLA uses target-region reconstruction to
shape task-relevant perception, while VLA-JEPA and DreamVLA supervise policies
with latent or structured future representations
\citep{reconvla,vlajepa,dreamvla}.

Related ideas also appear in video and world modeling. WorldVLA and DriveVLA-W0
use future visual generation, Motus makes motion-oriented dynamics explicit,
and VideoJAM, DreamWorld, and VideoREPA introduce motion, geometry, semantics,
or foundation-model features beyond RGB reconstruction
\citep{worldvla,drivevla-w0,motus,videojam,dreamworld,videorepa}. DreamWAM
brings these directions into a coupled VideoDiT--ActionDiT WAM: RGB and optical
flow are modeled through joint latent denoising, while heterogeneous geometry
and semantic features enter through gated residual branches to limit
interference with the pretrained video pathway. All beyond-RGB pathways are used only during training and removed during deployment.

\section{Method}
\label{sec:method}

\begin{figure*}[t]
    \centering
    \includegraphics[
        width=0.97\textwidth,
    ]{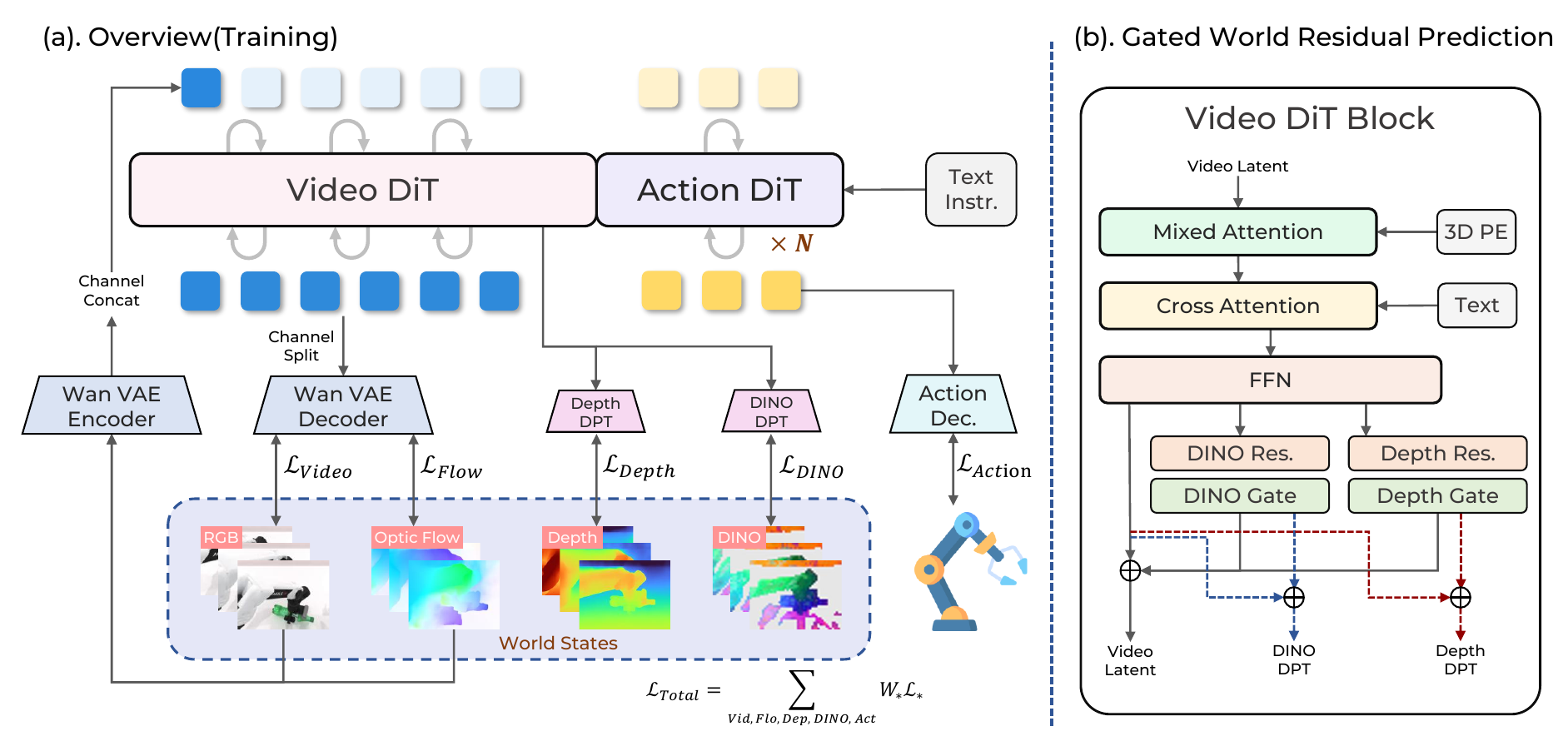}
    \caption{Overview of DreamWAM. (a) During training, DreamWAM augments RGB
    future prediction with motion, geometry, and semantic views. RGB and
    optical-flow latents are jointly denoised, while geometry and semantic
    targets shape selected VideoDiT layers through gated residual branches.
    Shared attention exposes the resulting representation to ActionDiT. All
    beyond-RGB supervision pathways are disabled at inference, which retains
    RGB video-action denoising. (b) Geometry- and semantic-oriented residual
    branches selectively update the VideoDiT representation during training.}
    \label{fig:dreamwam_method}
\end{figure*}

\subsection{Preliminaries}
\label{sec:preliminaries}

Given a language instruction $c$, a current observation $o_0$, future
observations $o_{1:H}$, and an action chunk $a_{1:N}$, a WAM jointly learns
future prediction and action generation. We build on the two-expert
VideoDiT-ActionDiT architecture of Fast-WAM~\citep{fastwam}: the VideoDiT
models future video latents, the ActionDiT denoises the action trajectory, and
paired layers exchange information through shared attention. A conventional
RGB-only WAM represents the future solely with $z^{\mathrm{rgb}}_{1:H}$.
DreamWAM retains the joint video-action denoising framework while extending the
future representation learned by the video expert beyond RGB.

\subsection{Structured Future Views}
\label{sec:future_views}

As shown in \autoref{fig:dreamwam_method}, DreamWAM represents the future as
\begin{equation}
    \mathcal{Y}_{1:H}
    =
    \left\{
        z^{\mathrm{rgb}}_{1:H},
        z^{\mathrm{mot}}_{1:H},
        f^{\mathrm{geo}}_{1:H},
        f^{\mathrm{sem}}_{1:H}
    \right\},
    \label{eq:future_views}
\end{equation}
where the four terms describe appearance, motion, geometry, and semantics,
respectively. These views provide complementary descriptions of future state
changes rather than a perfectly disentangled or complete physical state.

\paragraph{Appearance.}
The future RGB sequence is encoded by the Wan2.2 video VAE~\citep{wan22},
producing the standard appearance-oriented future latent
$z^{\mathrm{rgb}}$.

\paragraph{Motion.}
RAFT~\citep{raft} estimates dense optical flow between adjacent future frames. The resulting
flow sequence is encoded by the same video VAE to produce
$z^{\mathrm{mot}}$ on the RGB latent grid. This view makes image-space temporal
change an explicit future target without claiming metric 3D scene flow.

\paragraph{Geometry.}
Depth Anything V3~\citep{depthanythingv3} extracts geometry-oriented
features from future frames.
After temporal and spatial alignment with the VideoDiT grid, these features
form $f^{\mathrm{geo}}$, which captures depth structure and relative spatial
organization without requiring a complete 3D reconstruction.

\paragraph{Semantics.}
DINOv2~\citep{dinov2} extracts patch-level features from the same future
frames. Their aligned
representation $f^{\mathrm{sem}}$ preserves object- and region-level
consistency rather than serving as a semantic segmentation label.

All three non-RGB views are constructed offline from the training videos. They
serve as training targets rather than additional policy observations, and their
supervision pathways are inactive during deployment.

\subsection{Heterogeneous Future Modeling}
\label{sec:future_modeling}

Because the four future views have different representational forms and
different compatibility with the pretrained video latent space, DreamWAM models
them through two corresponding pathways.

\paragraph{Joint latent denoising for RGB and motion.}
RGB and motion share the same dense spatiotemporal structure. During training, DreamWAM therefore perturbs them at the same flow-matching
timestep~\citep{flowmatching} and concatenates them along the latent channel
dimension:
\begin{equation}
\begin{aligned}
    x_t^q &= (1-t)z^q+t\epsilon^q,
    \quad q\in\{\mathrm{rgb},\mathrm{mot}\},\\
    x_t^{\mathrm{joint}}
    &= \operatorname{Concat}_{C}
       \left[x_t^{\mathrm{rgb}},x_t^{\mathrm{mot}}\right].
\end{aligned}
\label{eq:joint_world_latent}
\end{equation}
The VideoDiT learns both views in a shared denoising stream. This preserves the
standard video-modeling pathway while making temporal change an explicit part
of the predicted future rather than leaving it implicit in RGB.

\paragraph{Gated residual modeling for geometry and semantics.}
Geometry and semantic features differ from VAE latents in both source and
representation. Concatenating them as additional denoising channels would
expose the entire VideoDiT stream to target statistics defined by external
encoders. Unlike RGB and flow latents, these features are not native to the
video VAE space; forcing the backbone to model them through the same state can
therefore interfere with the video-generation prior inherited from pretraining.
DreamWAM instead uses lightweight residual branches during training to shape
selected VideoDiT layers while retaining the original block output as the main
stream. For a
feature view $j\in\{\mathrm{geo},\mathrm{sem}\}$, the update at layer $\ell$ is
\begin{equation}
\begin{aligned}
    \bar h_\ell &= B_\ell(h_{\ell-1}),\\
    h_\ell &= \bar h_\ell
    + \sum_j g_\ell^j(\bar h_\ell,c)
      \odot R_\ell^j(\bar h_\ell),\\
    \widehat f^j &= P_j(h_L),
\end{aligned}
\label{eq:gated_residual}
\end{equation}
where $R_\ell^j$ produces a view-specific residual, $g_\ell^j$ controls its
magnitude, and $P_j$ predicts the aligned future feature during training. The
additive path keeps $\bar h_\ell$ directly available as the backbone state,
while the learned gate controls the magnitude of each view-specific correction.
Depth- and DINO-derived supervision therefore refines, rather than replaces,
the pretrained video representation. This design limits interference with the
native video pathway while still allowing action-relevant structure to enter
the shared VideoDiT state. It also avoids an independent denoising stream for
every future view. Accordingly, during training, VAE-aligned motion enters the full denoising
state, whereas external geometry and semantic targets influence the backbone
through controlled residual corrections. All three beyond-RGB supervision
signals are inactive at deployment.

\subsection{World-Action Coupling}
\label{sec:world_action_coupling}

DreamWAM uses a two-expert VideoDiT-ActionDiT architecture. The VideoDiT models
the future-state representation, while the ActionDiT denoises the action
trajectory. The experts retain separate parameters but exchange information
through shared attention at paired layers. Video and action tokens contribute
to the same attention context before each expert applies its own output
projection and feed-forward transformation. Consequently, the ActionDiT can
condition on the multi-view future representation learned by the VideoDiT
during training without directly consuming optical-flow, depth, or semantic
teacher features.

This interaction connects future modeling to the policy objective. The
additional future views are not optimized as isolated auxiliary tasks; they
reshape the VideoDiT representation that participates in action denoising,
allowing structured future prediction to directly influence action learning.

\subsection{Training Objective and Inference}
\label{sec:training_inference}

The overall objective combines dense future denoising, feature prediction,
action flow matching, and gate regularization:
\begin{equation}
\begin{split}
    \mathcal{L} ={}&
      \lambda_{\mathrm{rgb}}\mathcal{L}_{\mathrm{rgb}}^{\mathrm{FM}}
      +\lambda_{\mathrm{mot}}\mathcal{L}_{\mathrm{mot}}^{\mathrm{FM}}
      +\lambda_{\mathrm{geo}}\mathcal{L}_{\mathrm{geo}}^{\mathrm{pred}}\\
      &+\lambda_{\mathrm{sem}}\mathcal{L}_{\mathrm{sem}}^{\mathrm{pred}}
      +\lambda_{\mathrm{act}}\mathcal{L}_{\mathrm{act}}^{\mathrm{FM}}
      +\lambda_g\mathcal{L}_{\mathrm{gate}}.
\end{split}
\label{eq:total_objective}
\end{equation}
The gate regularizer $\mathcal{L}_{\mathrm{gate}}$ is an $\ell_1$ penalty on
the gate activations. All world targets are derived from future frames in the
corresponding training split. The motion pathway, gated residual branches, and feature-prediction heads
are active only during training, where their objectives update the shared
VideoDiT backbone.

At inference, DreamWAM follows the RGB-only deployment interfaces of
Fast-WAM. In the main configuration, the VideoDiT denoises the RGB latent while
the ActionDiT denoises the action chunk through shared attention. The motion
input is inactive; the residual branches, feature-prediction heads, RAFT, Depth
Anything V3, and DINOv2 are removed; and the precomputed training targets are
not used. The final control output is the denoised action chunk, with no
auxiliary representation prediction or online teacher inference.

\section{Experiments}
\label{sec:experiments}

We evaluate DreamWAM on standard and distribution-shifted manipulation tasks in
simulation and on a real robot, followed by controlled component and routing
ablations.

\subsection{Experimental Setup}
\label{sec:experimental_setup}

\paragraph{Simulation (LIBERO and LIBERO-Plus).}
LIBERO contains four 10-task suites---Spatial, Object, Goal, and Long---with 500
demonstrations per suite~\citep{libero}. We train only on the original
LIBERO demonstrations and evaluate 50 rollouts per task (2,000 per seed).
LIBERO-Plus evaluates the same skills under seven unseen shifts: camera
viewpoint, robot appearance, language, lighting, background, image noise, and
object layout~\citep{liberoplus}; no LIBERO-Plus data are used for training.
Because the task semantics and success criteria remain unchanged, the benchmark
isolates robustness to observation, instruction, and scene changes rather than
adaptation to new manipulation skills. Each seed is evaluated on 10,030
episodes. We average within each perturbation dimension and report the
unweighted mean of the seven dimension scores. Following Fast-WAM, we evaluate
two inference modes. Fast-WAM denotes its primary no-video-rollout setting,
which retains only the first-frame video input and denoises actions directly,
whereas Fast-WAM-Joint jointly denoises future RGB latents and actions.
DreamWAM is built primarily on the joint variant, enriching its learned future
with RGB, motion, geometry, and semantic views; DreamWAM-uncond evaluates the
learned model without video rollout and retains only the first-frame video
input. All matched variants share the Wan2.2-5B VideoDiT~\citep{wan22},
ActionDiT, training data, and a 32-step action policy that executes 10 steps
before replanning with 10 denoising steps. All simulation results from our implementations, including the ablations,
average two independent random seeds. We train all variants on eight NVIDIA
H20 GPUs using bfloat16, a learning rate of $1\times10^{-5}$, and a batch size
of 16. Future targets are precomputed. At inference, the optical-flow channels are filled with zeros and
are therefore inactive, while the residual and target-encoder pathways are
removed. Other methods are included only for benchmark context because their backbones, pretraining, and
implementations differ.

\paragraph{Real robot.}
We evaluate DreamWAM on an AgileX PiPER dual-arm platform, with 30 trials for
each task or perturbation setting. For all real-robot experiments, we adopt the
Fast-WAM-Joint setting and use it as the matched RGB-only baseline.
The standard evaluation contains four multi-step tabletop tasks:
\emph{Dish Stack} sequentially stacks two small plates on a large plate;
\emph{Dual-Object Pick} places a red cup and a black pen holder into a basket,
starting from the rightmost object; \emph{Strawberry Selection} identifies the
strawberries and places them on a plate; and \emph{Block Stack} moves two
colored blocks to the table center and stacks the orange block on the red one.
For robustness, we retain the Strawberry Selection instruction, target objects,
and success criterion while independently perturbing the scene in three ways:
random colored patches change the tabletop background, a fill light changes the
illumination, and task-irrelevant objects are added around the workspace as
visual distractors. No demonstrations from these perturbed settings are used
for training.

\begin{table}[t]
    \centering
    {\small
    \setlength{\tabcolsep}{3.5pt}
    \begin{tabular}{@{}lrrrrr@{}}
        \toprule
        Method & Spatial & Object & Goal & Long & Avg. \\
        \midrule
        OpenVLA~\citeyearpar{openvla}
        & 84.70 & 88.40 & 79.20 & 53.70 & 76.50 \\

        $\pi_0$~\citeyearpar{pi0}
        & 96.80 & 98.80 & 95.80 & 85.20 & 94.15 \\

        $\pi_{0.5}$~\citeyearpar{pi05}
        & 98.80 & 98.20 & 98.00 & 92.40 & 96.85 \\

        \midrule
        Motus~\citeyearpar{motus}
        & 96.80 & 99.80 & 96.60 & 97.60 & 97.70 \\

        LingBot-VA~\citeyearpar{lingbot-va}
        & 98.50 & 99.60 & 97.20 & 98.50 & 98.45 \\

        \midrule
        Fast-WAM~\citeyearpar{fastwam}
        & 97.00 & \underline{99.80} & 97.60 & 94.80 & 97.30 \\

        \textbf{DreamWAM-uncond}
        & \underline{99.20} & 99.60 & \underline{98.40}
        & \underline{96.40} & \underline{98.40} \\

        Fast-WAM-Joint~\citeyearpar{fastwam}
        & 99.40 & 98.20 & \textbf{98.80} & 95.60 & 98.00 \\

        \textbf{DreamWAM}
        & \textbf{99.60} & \textbf{99.80} & 98.60
        & \textbf{97.60} & \textbf{98.90} \\
        \bottomrule
    \end{tabular}
    }
    \caption{Success rates (\%) on LIBERO. Fast-WAM-Joint and DreamWAM are
    matched two-seed averages with 2,000 rollouts per seed; the no-rollout rows
    use the same task-level protocol. Underlining and boldface mark the better
    result within the no-rollout and joint settings, respectively.}
    \label{tab:libero_main}
\end{table}

\begin{table*}[t]
    \centering
    {\small
    \setlength{\tabcolsep}{3pt}
    \begin{tabular}{clcrrrrrrrr}
        \toprule
        Type & Method & Emb. PT
        & Cam. & Robot & Lang. & Light & Bg. & Noise & Layout & Avg. \\
        \midrule
        \multirow{4}{*}{VLA}
        & UniVLA~\citep{univla} & $\checkmark$
        & 1.80 & 46.20 & 69.60 & 69.00
        & 81.00 & 21.20 & 31.90 & 45.81 \\

        & OpenVLA-OFT~\citep{openvla-oft} & $\checkmark$
        & \underline{56.40} & 31.90 & 79.50 & 88.70
        & \textbf{93.30} & \underline{75.80} & 74.20
        & \underline{71.40} \\

        & $\pi_0$~\citep{pi0} & $\checkmark$
        & 13.80 & 6.00 & 58.80 & 85.00
        & \underline{81.40} & \textbf{79.00} & 68.90 & 56.13 \\

        & $\pi_0$-FAST~\citep{pi0-fast} & $\checkmark$
        & \textbf{65.10} & 21.60 & 61.00 & 73.20
        & 73.20 & 74.40 & 68.80 & 62.47 \\

        \midrule
        \multirow{5}{*}{WAM}
        & WorldVLA~\citep{worldvla} & $\checkmark$
        & 0.10 & 27.90 & 41.60 & 43.70
        & 17.10 & 10.90 & 38.00 & 25.61 \\

        & Fast-WAM~\citep{fastwam} & $\times$
        & 16.26 & 44.13 & 66.82 & 79.77
        & 52.60 & 38.54 & 61.38 & 51.36 \\

        & \textbf{DreamWAM-uncond} & $\times$
        & 30.39 & 59.23 & 77.29 & 85.90
        & 64.68 & 56.65 & 69.97 & 63.44 \\

        & Fast-WAM-Joint~\citep{fastwam} & $\times$
        & 39.59 & \underline{60.90} & \underline{92.32}
        & \underline{94.57} & 57.62 & 58.59
        & \underline{80.52} & 69.16 \\

        & \textbf{DreamWAM} & $\times$
        & 53.78 & \textbf{63.61} & \textbf{94.80}
        & \textbf{96.67} & 71.56 & 67.15
        & \textbf{80.72} & \textbf{75.47} \\
        \bottomrule
    \end{tabular}
    }
    \caption{Success rates (\%) under LIBERO-Plus perturbations. Avg. is the
    unweighted mean of the seven columns. Fast-WAM and DreamWAM-uncond use no
    video rollout, whereas the joint variants denoise future RGB and actions.
    Fast-WAM-Joint and DreamWAM average two seeds, with 10,030 episodes per seed;
    no LIBERO-Plus data are used for training. Bold and underlined values denote
    the best and second-best results.}
    \label{tab:libero_plus}
    \vspace{-5mm}
\end{table*}

\subsection{Results}
\label{sec:main_results}

\subsubsection{LIBERO}
\label{sec:libero_results}

\autoref{tab:libero_main} shows consistent gains under both inference modes:
DreamWAM-uncond improves Fast-WAM from 97.30\% to 98.40\%, while DreamWAM
improves Fast-WAM-Joint from 98.00\% to 98.90\%. DreamWAM consequently
achieves the highest overall average among the compared methods, while
DreamWAM-uncond remains competitive with the strongest prior methods despite
omitting video rollout. The no-rollout gain shows that structured future supervision strengthens the policy itself, while the stronger joint variants indicate that such supervision
and test-time imagination are complementary.

\subsubsection{LIBERO-Plus}
\label{sec:libero_plus_results}

It evaluates out-of-distribution generalization while preserving the
underlying manipulation skills. As shown in \autoref{tab:libero_plus}, DreamWAM
improves all seven perturbation dimensions in both inference modes. The
no-rollout setting provides the key contrast: the average rises from 51.36\% to
63.44\% even without test-time imagination, while joint inference also improves
from 69.16\% to 75.47\%. Since each matched pair shares the backbone, training
data, and evaluation protocol, the gains indicate that structured future
supervision is internalized by the policy and remains effective when future RGB
imagination is retained.

DreamWAM also attains the highest perturbation average among the compared
methods without embodied pretraining, although heterogeneous backbones and
training data make the matched Fast-WAM pairs the primary controlled evidence.
The two inference modes separate learning a structured future from generating
one at test time: structured supervision improves the no-rollout policy by
12.09 points, while joint rollout remains beneficial. The reduced
no-rollout-to-joint gap after DreamWAM training (12.02 versus 17.80 points)
suggests that part of the action-relevant future structure otherwise supplied
online has already been internalized. Training-time future representation and
test-time imagination are therefore complementary rather than interchangeable.

\begin{figure*}[t]
    \centering
    \includegraphics[width=0.97\textwidth]{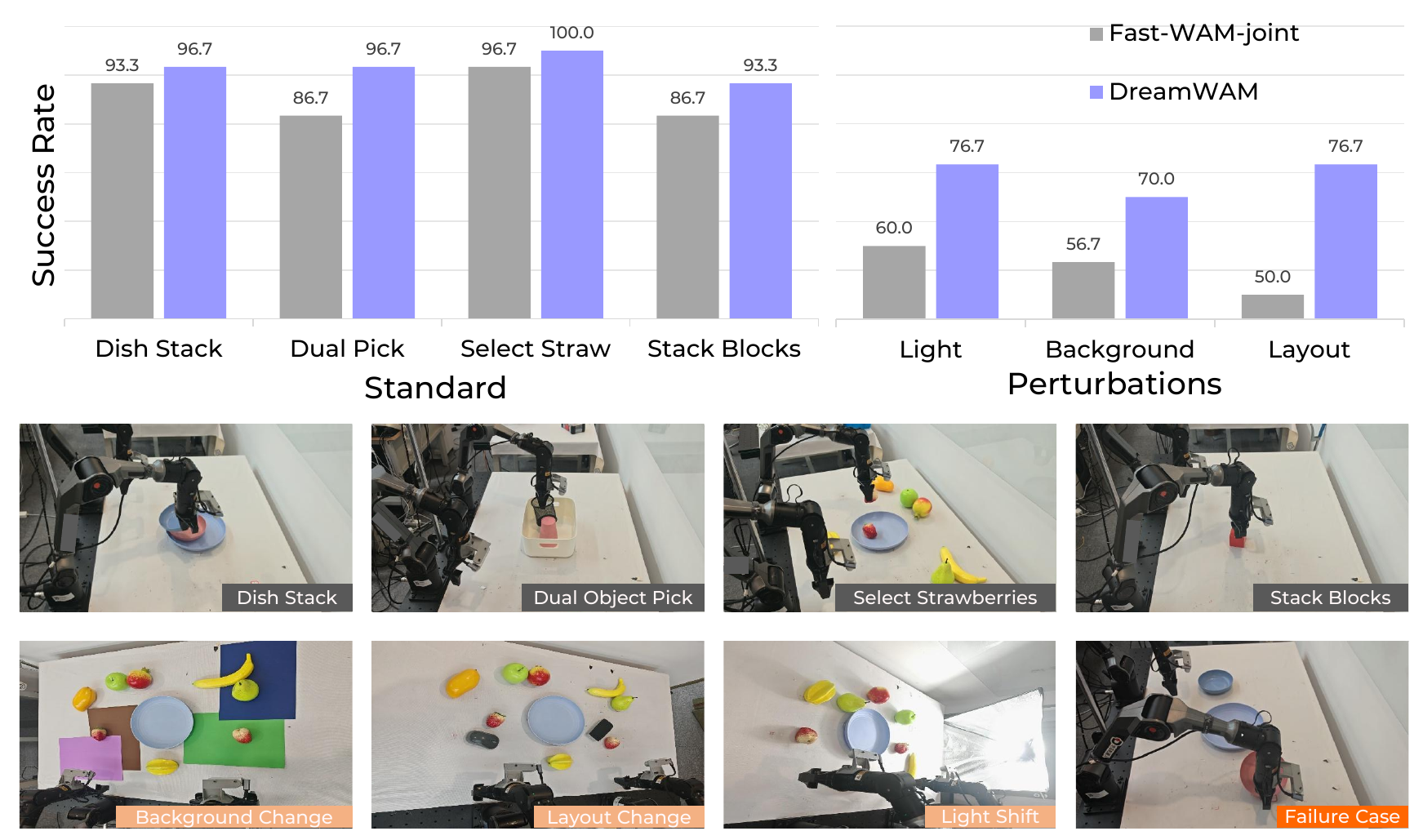}
    \caption{\textbf{Real-world robot evaluation.} Top: success rates (\%) on
    four standard tasks and three unseen visual perturbations, with 30 trials
    per setting. Bottom: representative task and perturbation observations. The
    perturbations retain the Strawberry Selection instruction and success
    criterion. The rightmost panel shows a Fast-WAM-Joint failure caused by an
    inaccurate gripper-to-plate spatial relation.}
    \label{fig:real_world_results}
\end{figure*}

\subsubsection{Real-World Robot}
\label{sec:real_world_results}

As shown in \autoref{fig:real_world_results}, DreamWAM improves all seven
real-world settings. The average rises from 90.8\% to 96.7\% on the standard
tasks and from 55.6\% to 74.4\% under unseen visual perturbations, making the
robustness gain substantially larger than the in-distribution gain. This mirrors
the LIBERO-Plus trend and indicates that structured future supervision improves
resilience to changes in visual realization rather than adaptation to new task
semantics.

Across simulation and real-world evaluation, DreamWAM yields modest gains in
the original environments but substantially larger gains when visual
realization changes while task semantics remain fixed. Because this pattern
recurs without rollout, with joint imagination, and on real hardware, it is more
consistent with preserving action-relevant state transitions than with uniformly
improving in-distribution fitting.

The highlighted failure provides a qualitative example of this difference.
Fast-WAM-Joint closes the gripper beside the plate, revealing an inaccurate
gripper-to-plate spatial relation before grasp execution rather than a failure
to understand the instruction. A single rollout does not isolate the cause by
itself, but the error is consistent with the aggregate perturbation results:
RGB-only future prediction need not explicitly preserve object displacement,
relative geometry, and task-relevant identity when the surrounding pixels
change. DreamWAM's motion, geometry, and semantic future supervision jointly
encourage the policy to retain these action-relevant relations.

\begin{table}[t]
    \centering
    {\small
    \setlength{\tabcolsep}{2.4pt}
    \sisetup{
        table-format=2.2,
        table-number-alignment=center,
        detect-weight=true,
        detect-inline-weight=math
    }

    \begin{tabular}{@{}lccc@{\hspace{5pt}}SSSSS@{}}
        \toprule
        Variant & Mot. & Geo. & Sem.
        & {Spa.} & {Obj.} & {Goal} & {Long} & {Avg.} \\
        \midrule

        RGB only
        & -- & -- & --
        & {\underline{99.40}}
        & 98.20
        & {\underline{98.80}}
        & 95.60
        & 98.00 \\

        \midrule

        Mot. only
        & $\mathrm{R}$ & -- & --
        & 99.00
        & 98.20
        & 98.60
        & 95.60
        & 97.85 \\

        Mot. only
        & $\mathrm{D}$ & -- & --
        & 98.60
        & {\bfseries 99.80}
        & 98.20
        & 97.40
        & 98.50 \\

        Geo. only
        & -- & $\mathrm{R}$ & --
        & {\underline{99.40}}
        & {\underline{99.60}}
        & 97.60
        & 96.00
        & 98.15 \\

        Geo. only
        & -- & $\mathrm{D}$ & --
        & 97.20
        & {\underline{99.60}}
        & 96.20
        & 94.20
        & 96.80 \\

        Sem. only
        & -- & -- & $\mathrm{R}$
        & 99.00
        & 97.40
        & {\bfseries 99.20}
        & 96.80
        & 98.10 \\

        Sem. only
        & -- & -- & $\mathrm{D}$
        & 96.80
        & 98.80
        & 97.40
        & 94.20
        & 96.80 \\

        \midrule

        w/o Mot.
        & -- & $\mathrm{R}$ & $\mathrm{R}$
        & 98.80
        & 99.20
        & 97.00
        & 95.00
        & 97.50 \\

        w/o Geo.
        & $\mathrm{D}$ & -- & $\mathrm{R}$
        & 98.40
        & 99.00
        & 98.40
        & 97.20
        & 98.25 \\

        w/o Sem.
        & $\mathrm{D}$ & $\mathrm{R}$ & --
        & 99.20
        & 99.00
        & 98.20
        & {\bfseries 98.00}
        & {\underline{98.60}} \\

        \midrule

        Denoise
        & $\mathrm{D}$ & $\mathrm{D}$ & $\mathrm{D}$
        & 98.60
        & 98.60
        & 97.80
        & 96.40
        & 97.85 \\

        \textbf{Hybrid}
        & $\mathrm{D}$ & $\mathrm{R}$ & $\mathrm{R}$
        & {\bfseries 99.60}
        & {\bfseries 99.80}
        & 98.60
        & {\underline{97.60}}
        & {\bfseries 98.90} \\

        \bottomrule
    \end{tabular}
    }

    \caption{Component and routing ablation on LIBERO. Mot., Geo., and Sem.
    denote motion, geometry, and semantics. All variants retain RGB future
    prediction. -- denotes no injection, $\mathrm{D}$ full denoising, and
    $\mathrm{R}$ gated residual injection. Denoise uses $\mathrm{D}/\mathrm{D}/
    \mathrm{D}$, whereas Hybrid is the proposed $\mathrm{D}/\mathrm{R}/
    \mathrm{R}$ routing. Bold and underlined values indicate the best and
    second-best results in each column; ties receive the same formatting.}
    \label{tab:target_ablation}
\end{table}

\subsection{Ablation Study}
\label{sec:target_analysis}

The ablation addresses two questions: whether the three beyond-RGB views
provide complementary supervision, and whether they should enter the VideoDiT
through the same pathway. \autoref{tab:target_ablation} compares the RGB-only
baseline, single-view and leave-one-view-out variants, an all-denoise variant,
and complete DreamWAM under
the same training protocol. All variants retain RGB future prediction.
$\mathrm{D}$ denotes full denoising through channel concatenation, and
$\mathrm{R}$ denotes gated residual injection. The paired single-view rows hold
the target fixed and vary only its route, while the all-denoise control keeps
all three targets fixed and changes the joint routing from
$\mathrm{D}/\mathrm{R}/\mathrm{R}$ to
$\mathrm{D}/\mathrm{D}/\mathrm{D}$.

\paragraph{Complementary future views.}
Under their preferred routes, all three auxiliary views improve the RGB-only
baseline. The leave-one-view-out comparison identifies motion as the most
consequential signal: removing it causes the largest degradation, while
geometry and semantics without motion underperform either feature alone. This
suggests that motion provides a temporal scaffold for integrating the two
feature-level signals, enabling the complete
$\mathrm{D}/\mathrm{R}/\mathrm{R}$ model to achieve the best overall balance.

\paragraph{Representation-matched injection.}
The paired controls show a consistent representation-matching effect: full
denoising is preferable for motion, whereas residual injection is preferable
for both geometry and semantics. The all-denoise control provides the key
contrast: despite using the same three targets,
$\mathrm{D}/\mathrm{D}/\mathrm{D}$ falls below the RGB-only baseline, while
$\mathrm{D}/\mathrm{R}/\mathrm{R}$ reaches 98.90\%. This indicates that the
gain comes from matching each target to a compatible route---VAE-aligned motion
in the denoising state and heterogeneous feature targets as controlled residual
corrections---rather than from supervision quantity alone. Together, these
controls show that adding auxiliary objectives alone is insufficient: both the
information carried by the future views and the route through which each view
enters the pretrained representation matter.

\section{Conclusion}
\label{sec:conclusion}

We introduced DreamWAM, which extends RGB-only future prediction with motion,
geometry, and semantic supervision in a coupled VideoDiT--ActionDiT WAM.
During training, RGB and optical-flow latents are jointly denoised, while
Depth- and DINO-derived targets are incorporated through gated residual
branches; inference follows the RGB-only Fast-WAM deployment interface without
online auxiliary prediction. Across no-rollout and joint inference, DreamWAM
improves LIBERO from 97.30\% to 98.40\% and from 98.00\% to 98.90\%, and
LIBERO-Plus from 51.36\% to 63.44\% and from 69.16\% to 75.47\%, respectively.
It also improves success under the evaluated real-world perturbations from
55.6\% to 74.4\%, providing initial evidence that the gains extend beyond
simulation. Controlled ablations further support full denoising for VAE-aligned motion and
residual injection for heterogeneous geometry and semantic features. Future work can explore temporally consistent metric 3D and contact-aware targets while retaining the training-time structured supervision and RGB-only deployment.

\bibliography{aaai2027}

\setcounter{secnumdepth}{2}
\appendix
\section{Implementation Details}
\label{app:implementation_details}

\subsection{Beyond-RGB Target Construction}
\label{app:target_construction}

All beyond-RGB targets are precomputed offline from temporally aligned
nine-frame RGB clips in the training set. Tensor dimensions below follow the
channel, time, height, and width order. The motion representation is encoded on
the same Wan2.2 VAE grid as the RGB video latent, whereas the geometry and
semantic targets are spatially and temporally aligned with this grid before
supervising the training-only residual branches.

\paragraph{RAFT motion latent.}
For each pair of adjacent frames in an aligned nine-frame clip, a frozen RAFT
model~\citep{raft} estimates a dense two-dimensional optical-flow field. The
eight resulting flow fields are converted into color-coded RGB flow
visualizations. We duplicate the first transition at the beginning of the
sequence so that the flow video also contains nine frames and remains aligned
with the RGB clip. The resulting flow video is encoded by the same frozen
Wan2.2 VAE~\citep{wan22} used for RGB, producing
\begin{equation}
    \mathbf{z}_{\mathrm{flow}}
    \in \mathbb{R}^{48\times3\times14\times28}.
    \label{eq:app_flow_latent}
\end{equation}
Thus, $\mathbf{z}_{\mathrm{flow}}$ is the VAE encoding of a visualized motion
sequence rather than a latent representation of raw metric flow vectors or 3D
scene flow.

\paragraph{DA3 geometry latent.}
Each RGB frame contains horizontally concatenated camera observations. We first
split these observations into their individual views and apply a frozen
DA3-Base model from Depth Anything V3~\citep{depthanythingv3} to obtain a depth
map for each view. The per-view depth maps are then restored to their original
horizontal layout. We apply a logarithmic transformation to the depth values
and partition each map into local regions aligned with the $14\times28$ spatial
grid of the video latent. The resulting local descriptors are projected to
eight channels using a rank-8 PCA basis fitted exclusively on training-set
depth features.

To align the nine frame-level features with the three-step video-latent
timeline, we retain the feature from the first frame, average the features from
frames two through five, and average those from frames six through nine. This
produces
\begin{equation}
    \mathbf{z}_{\mathrm{depth}}
    \in \mathbb{R}^{8\times3\times14\times28}.
    \label{eq:app_depth_latent}
\end{equation}

\paragraph{DINOv2 semantic latent.}
For every RGB frame, we extract normalized patch tokens using a frozen DINOv2
ViT-B/14 model with register tokens~\citep{dinov2}. The resulting
768-dimensional patch descriptors are projected to eight dimensions using a
PCA basis fitted only on training-domain features. The projected feature maps
are resized to the $14\times28$ video-latent grid and temporally aligned using
the same three-step aggregation as the depth features: the first frame is
retained, frames two through five are averaged, and frames six through nine are
averaged. Channel-wise normalization is applied after alignment, yielding
\begin{equation}
    \mathbf{z}_{\mathrm{DINO}}
    \in \mathbb{R}^{8\times3\times14\times28}.
    \label{eq:app_dino_latent}
\end{equation}
This representation serves as a self-supervised semantic target rather than a
set of class labels or segmentation masks.

\subsection{Zero-Filled Flow Channels at Inference}
\label{app:zero_flow_inference}

RAFT is not executed during deployment. After the current RGB observation is
encoded as $\mathbf{z}_{\mathrm{RGB}}^{0}$, we allocate an all-zero flow latent
with 48 channels and the same spatiotemporal resolution. The input to the
VideoDiT is constructed as
\begin{equation}
    \mathbf{x}_{\mathrm{infer}}
    = \operatorname{Concat}_{C}
      \left[
        \mathbf{z}_{\mathrm{RGB}}^{0},
        \mathbf{0}_{\mathrm{flow}}
      \right].
    \label{eq:app_inference_input}
\end{equation}
Partition the input projection according to the RGB and flow channel groups as
\begin{equation}
    \mathbf{W}
    = [\mathbf{W}_{\mathrm{RGB}},\mathbf{W}_{\mathrm{flow}}].
    \label{eq:app_partitioned_projection}
\end{equation}
The projected input then becomes
\begin{equation}
\begin{aligned}
    \mathbf{W}\mathbf{x}_{\mathrm{infer}} + \mathbf{b}
    &= \mathbf{W}_{\mathrm{RGB}}
       \mathbf{z}_{\mathrm{RGB}}^{0} \\
    &\quad + \mathbf{W}_{\mathrm{flow}}
       \mathbf{0}_{\mathrm{flow}} + \mathbf{b} \\
    &= \mathbf{W}_{\mathrm{RGB}}
       \mathbf{z}_{\mathrm{RGB}}^{0} + \mathbf{b}.
\end{aligned}
\label{eq:app_zero_flow_projection}
\end{equation}
The flow channels therefore make no direct contribution to the VideoDiT input
at inference, while the shared VideoDiT parameters retain the representation
learned from joint RGB--motion supervision. Training partially matches this
boundary condition: the flow channels aligned with the conditioning RGB frame
are also set to zero, and the flow-denoising objective is applied only to
future latent steps. The RAFT, DA3, and DINOv2 encoders, together with the
geometry and semantic residual branches and their prediction heads, are not
executed during deployment.

\section{Real-World Experimental Details}
\label{app:real_world_details}

\subsection{Training and Evaluation Protocol}
\label{app:real_world_protocol}

The real-world training set contains 400 demonstration rollouts for each of the
four tasks, giving 1,600 rollouts in total. Every rollout is paired with its
corresponding natural-language task description. We convert all demonstrations
to the LeRobot v2.1 format, pool the four task datasets into a single multi-task
training set, and train each model for five epochs. Fast-WAM-Joint~\citep{fastwam}
and DreamWAM use the same pooled training data and training schedule.

After training, each model is evaluated over 30 rollouts for each of the four
standard tasks. The three visual perturbations are evaluated separately on the
Strawberry Selection task, with 30 rollouts per model for each perturbation.
Accordingly, each model is evaluated over 210 real-world rollouts in total. The
success rate for an individual task or perturbation setting is computed as
\begin{equation}
    \mathrm{Success\ Rate}
    = \frac{N_{\mathrm{success}}}{30}\times100\%.
    \label{eq:app_real_success_rate}
\end{equation}
The reported standard-task average is the arithmetic mean over the four task
success rates, and the perturbation average is the arithmetic mean over the
three perturbation success rates. The same trained model is evaluated in all
three perturbed settings without additional adaptation, and no demonstrations
from these settings are used for training.

\subsection{Task Descriptions}
\label{app:real_world_tasks}

\paragraph{Dual-Object Pick.}
The internal task identifier is \texttt{dual\_box\_\allowbreak pick}. The robot clears the tabletop by placing the red cup and the black pen holder
into the basket. It starts with the item initially located on the right and then
transfers the remaining item.

\paragraph{Strawberry Selection.}
The internal task identifier is \texttt{select\_\allowbreak strawberry}. The robot identifies the strawberries among the other fruits and places them on
the plate.

\paragraph{Block Stack.}
The internal task identifier is \texttt{Stack\_two\_\allowbreak blocks}. The robot moves the red and orange blocks to the center of the table, uses the
red block as the base, and stacks the orange block on top.

\paragraph{Dish Stack.}
The internal task identifier is \texttt{Stack\_dish}. The robot first places the small red plate on the large plate and then stacks
the small blue plate on top of the red plate.

%

\subsection{Visual Perturbations}
\label{app:real_world_perturbations}

The three perturbations are applied separately to the Strawberry Selection
task. In every setting, the language instruction, target fruits, plate, and
required manipulation remain unchanged, so the evaluation isolates a single
visual distribution shift at a time.

\paragraph{Background.}
Randomly positioned colored patches are introduced into the visible tabletop
background. They alter the local colors and textures of the scene without
changing the target objects or the required action.

\paragraph{Lighting.}
A fill light is placed at different angles and operated at different
intensities to change the illumination direction, scene brightness, shading,
and reflections while leaving the task itself unchanged.

\paragraph{Distractor Objects.}
Objects unrelated to the task are randomly placed on the tabletop around the
workspace. These additional objects introduce visual clutter and competing
object appearances but do not change which fruits are targets or where the
strawberries must be placed.

%

\subsection{Success Criterion}
\label{app:real_world_success}

For every standard task and perturbation setting, a rollout is counted as
successful only if the robot completes the action sequence specified by the
corresponding task description and all manipulated objects remain stably at
their required final locations when the rollout ends. Any other outcome is
counted as a failure, and no partial credit is assigned.

\end{document}